\documentclass[10pt,twocolumn,letterpaper]{article}

\usepackage{iccv}
\usepackage{times}
\usepackage{epsfig}
\usepackage{graphicx}
\usepackage{amsmath}
\usepackage{amssymb}
\usepackage{float}
\usepackage{caption}

\usepackage{graphicx}
\usepackage{amsmath}
\usepackage{amssymb}
\usepackage{booktabs}

\usepackage{bm}
\usepackage{amsfonts}
\usepackage{algorithmic}
\usepackage{textcomp}

\usepackage{algorithm}% http://ctan.org/pkg/algorithms
\usepackage{pifont} % \ding{51}
\usepackage{bbding} % \XSolidBrush
\usepackage{bbm}
\usepackage{indentfirst}

\usepackage{multirow}
\usepackage[T1]{fontenc}
\usepackage{color}

\usepackage[breaklinks=true,bookmarks=false]{hyperref}

\title{Towards Fine-grained Interactive Segmentation in Images and Videos}
\author{
    Yuan Yao \quad
    Qiushi Yang \quad
    Miaomiao Cui \quad
    Liefeng Bo \\[5pt]
    Institute for Intelligent Computing, Alibaba Group \\
    {\tt\small \{ryan.yy, yangqiushi.yqs, miaomiao.cmm, liefeng.bo\}@alibaba-inc.com} 
}

\begin{document}

\twocolumn[{
\renewcommand\twocolumn[1][]{#1}
\maketitle
\begin{center}
    \vspace{-20pt}
    \includegraphics[width=1.0\linewidth]{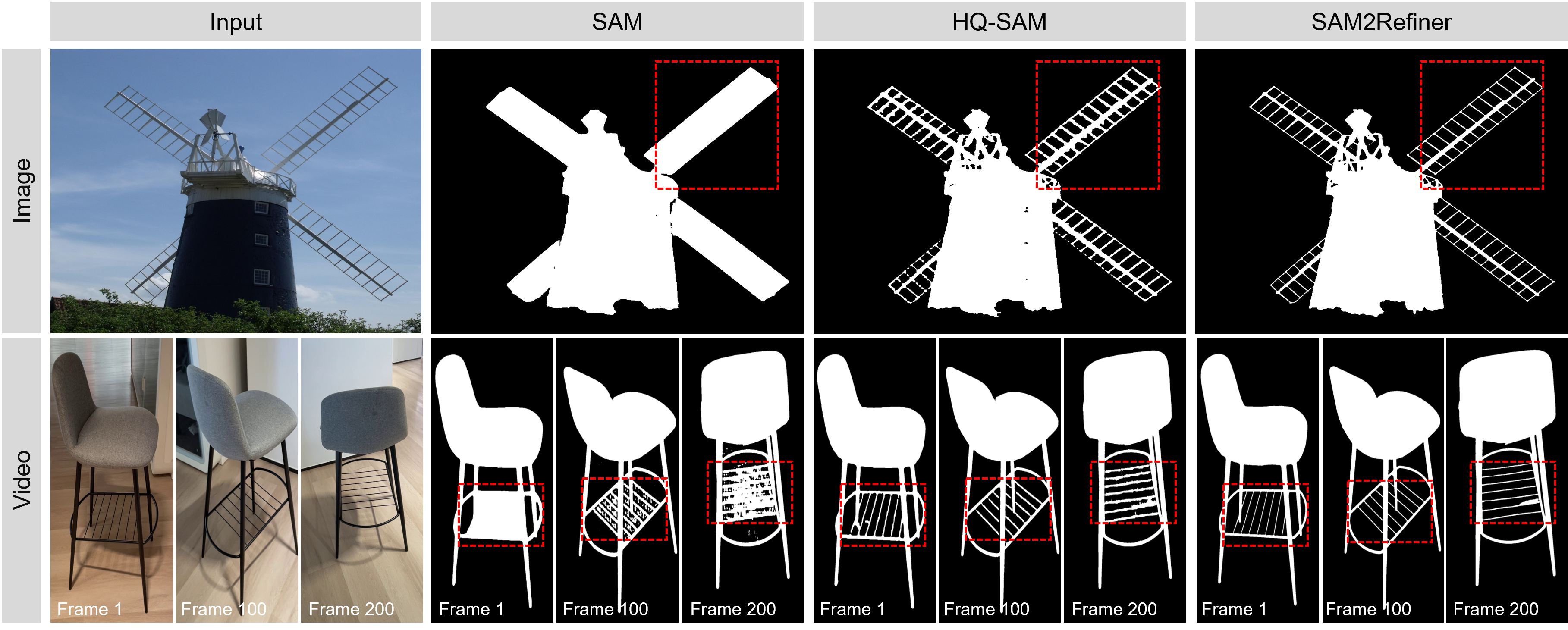}
    \vspace{-20pt}
    \captionsetup{type=figure}
    \caption{Comparison of segmentation results of the proposed framework SAM2Refiner and previous methods. The proposed SAM2Refiner achieves the best performance on both image and video segmentation, especially in fine-grained details.}
    \label{fig:visualization1}
    \vspace{20pt}
\end{center}
}]
%-------------------------------------------------------------------------
% \begin{figure*}
% \begin{center}
% %\setlength{\abovecaptionskip}{0cm}
% \setlength{\belowcaptionskip}{-0.1cm}
% \includegraphics[width=0.98\linewidth]{pic/architecture.png}
% \caption{
% An overview of the proposed framework. 
% } 
% \label{fig:network}
% \end{center}
% \end{figure*}

% \ificcvfinal\thispagestyle{empty}\fi
%-------------------------------------------------------------------------

\begin{abstract}
The recent Segment Anything Models (SAMs) have emerged as foundational visual models for general interactive segmentation. 
Despite demonstrating robust generalization abilities, they still suffer performance degradations in scenarios demanding accurate masks. 
Existing methods for high-precision interactive segmentation face a trade-off between the ability to perceive intricate local details and maintaining stable prompting capability, which hinders the applicability and effectiveness of foundational segmentation models.
To this end, we present an SAM2Refiner framework built upon the SAM2 backbone. 
This architecture allows SAM2 to generate fine-grained segmentation masks for both images and videos while preserving its inherent strengths. 
Specifically, we design a localization augment module, which incorporates local contextual cues to enhance global features via a cross-attention mechanism, thereby exploiting potential detailed patterns and maintaining semantic information.
% and further employs a UNet-like structure to fuse mask features with hierarchical features from the encoder. 
Moreover, to strengthen the prompting ability toward the enhanced object embedding, we introduce a prompt retargeting module to renew the embedding with spatially aligned prompt features. 
In addition, to obtain accurate high resolution segmentation masks, a mask refinement module is devised by employing a multi-scale cascaded structure to fuse mask features with hierarchical representations from the encoder.
Extensive experiments demonstrate the effectiveness of our approach, revealing that the proposed method can produce highly precise masks for both images and videos, surpassing state-of-the-art methods.

\end{abstract}
%-------------------------------------------------------------------------

\vspace{-2pt}
\section{Introduction}
\vspace{-2pt}
\label{sec:intro}

Interactive segmentation is a long-standing computer vision task that allows users to segment objects
from the guidance of user interaction, typically in the form of clicks, strokes, or bounding boxes. In recent years,
the Segment Anything Model (SAM)~\cite{sam} has emerged as a foundational visual model for general
interactive segmentation. Trained on their proposed SA-1B dataset, which consists of more than 1 billion
masks and 11 million images, SAM can generate masks in various scenes and has been widely applied to
a wide range of downstream vision tasks, including image editing~\cite{xie2023edit,yu2023inpaint}, 3D vision
~\cite{shen2023anything,xu2024embodiedsam} and medical image analysis~\cite{ma2024segment,wu2023medical}. The following work Segment Anything Model 2 (SAM2)~\cite{ravi2024sam} further introduces the streaming memory mechanism to extend the ability of SAM to videos, providing a potential boost to video applications such as AR / VR, robotics and autonomous vehicles.

Despite demonstrating robust generalization abilities, SAMs encounter the problems in scenario demanding fine-grained segmentation. In particular, as shown in Figure~\ref{fig:visualization1}, they struggle to predict a complete mask for objects with fine structures or sharp boundaries, leading to overly smooth results or coarse boundaries. Thus, these types of failure may hinder the applicability of SAMs. 
Valuable efforts~\cite{hq-sam,liu2024segment,ma2024segment,ravi2024sam} have been devoted to solve these challenges by introducing extra input tokens and explicitly fusing the multi-level features; however, they are still limited by SAM's ability, thus failing to achieve satisfactory fine-grained segmentation.
To this end, we aim to start with a holistic analysis on the three key points of fine-grained segmentation toward mining detailed cues and global semantic information guided by the user prompts. 
Firstly, considering that the whole input image contains global information of semantics while it lacks the detailed representations on details, we are expected to dynamically harvest the complementary local and global features within the input image. 
Secondly, SAMs directly combine the prompt with the object features via the decoder, which fail to build strong correlation among prompts with fine-grained cues and are hard to activate the response on object details, producing biased segmentation. Hence, strengthening the correspondences between prompt features and object details is crucial to yield a high response intensity of prompts on local details, which is beneficial to fine-grained segmentation.
Thirdly, the object features with low resolution projected by the SAM's encoder may lose much detailed cues, to recover the missed fine-grained local patterns for yielding high-resolution segmentation maps, the multi-scale features within SAMs' encoder are important to be re-used.
% HQ-SAM~\cite{hq-sam} introduces a learnable hq-output token into SAM's mask decoder and operates on a fused feature to improve the segmentation precision of SAM. Moreover, the lightweight design allows flexible equipment with SAM2 by simply replace the updated decoder, achieving equal refinement for video tasks. Although effective yet efficient, HQ-SAM is fundamentally undertaking an error correction process, its performance is still limited by SAM's ability to handle fine-grained local details, thus failing to achieve satisfactory segmentation (see Figure 1). 
A cocurrent research Pi-SAM~\cite{liu2024segment} is the most similar work to urs, which proposes an additional mask decoder to expand the capability of SAM in high-resolution images. Nevertheless, it compromises with diminished responsiveness to visual prompt, necessitating a complementary module for error correction, and exhibit biased preferences for background clicks. Differently, our goal is to facilitate prompt response and consider both local and global presentations for efficient fine-grained segmentation.

% Pi-SAM~\cite{liu2024segment} proposes an additional mask decoder to expands SAM’s ability at high resolution images, However, the proposed encoder tends to compromise with diminished responsiveness to visual prompt, necessitating a complementary module for error correction. The proposed precise interactor, which needs to train independently, may exhibit biased preferences for background clicks. Overall, current methods are either fall short in their ability to handle fine-grained local details, or sacrifice the flexibility of the prompting ability.

To achieve the aforementioned goals, we propose the SAM2Refiner architecture, which includes three components for fine-grained interactive segmentation. 
% SAM2Refiner possesses a strong perception for both global semantic and local details, while providing stable prompting response for intricate objects and high-resolution images.
Specifically, we introduce Localization Augment (LA) to refine the details of global features with local contextual feature cues. 
% As mentioned in DIS task~\cite{yu2024multi}, features extracted with global receptive field in transformers-based methods may not handle fine-grained local details as good as CNNs. 
We split the input image into four sub-images, and send to SAM2's image encoder together with raw image. A global-local cross-attention mechanism is designed to highlight the local information about the object within the global representation among deep features. 
Moreover, to improve the response intensity of prompting for segmenting objects with intricate structures, we propose a Prompt Retargeting (PR) strategy by providing precise alignment between the prompts and enhanced object feature. PR performs spatial alignment between prompt features and object features, and then incorporates them to yield an enhanced prompt response.
% The key design is to replace the dense features of SAM2's prompt encoder to features extracted from dense prompt maps, and updates the object features with spatial aligned prompt features. 
In addition, to generate higher resolution masks with finer details, the Mask Refinement (MR) module is devised by incorporating the object features and multi-scale features from the original image. In MR, the intermediate features within the SAM2 encoder is re-used to fully preserve the original zero-shot capabilities.
% SAM2Refiner is integrated with SAM2 backbone and re-uses its intermediate features, thereby fully preserving the zero-shot capabilities. 
Although designed for image segmentation, SAM2Refiner exhibits strong compatibility with the video streaming pipeline (as in Figure~\ref{fig:visualization1}), allowing fine-grained segmentation for both images and videos. 
% To evaluate the effectiveness of SAM2Refiner, we train our model on HQSeg-44K dataset~\cite{ke2024segment}, and assess it on a suite of segmentation benchmarks, including both image and video segmentations. 
Extensive experiments on both image and video segmentation benchmarks demonstrate the effectiveness of our approach over state-of-the-art methods.

Our main contributions can be summarized as follows.
% \vspace{-2pt}
% itemize
\begin{itemize}
\item We propose SAM2Refiner, a new architecture capable of generating fine-grained segmentation masks with strong perception for both global semantic and local details, and simultaneously maintaining stable prompting ability and powerful zero-shot capability.
% \vspace{-2pt}
\item In SAM2Refiner, the localization augment is designed to perform holistic representation for inputs. It incorporates sub-images as additional input, thereby enhancing global features with object localization.
% and utilizes multi-scale feature fusion to refine the predicted masks.
% \vspace{-2pt}
\item To boost the response ability of prompting in intricate regions, we propose a prompt retargeting module by providing precise alignment between spatial-aware dense prompt features and the enhanced object feature.
% \vspace{-2pt}
\item Toward producing high-resolution segmentation maps with fine details, the mask refinement module is crafted to integrate multi-scale representations with object features.
% \vspace{-2pt}
\item Building upon the SAM2 backbone, our architecture exhibits strong compatibility to the video streaming pipeline, exhibiting state-of-the-art fine-grained segmentation on both images and videos benchmarks.
% \vspace{-2pt}
\end{itemize}

%-------------------------------------------------------------------------

\section{Related Work}
\label{sec:related}

\subsection{Interactive Segmentation}
Interactive segmentation enables users to provide cues for target regions and guide the segmentation process. User interactions can take various forms, including scribbles~\cite{sam,sam2,hq-sam,SegGPT}, bounding boxes~\cite{sam,hq-sam,liu2024segment}, clicks~\cite{yuan2020segfix,sam}, or language prompts~\cite{SegGPT,SEEM}. Traditional methods utilize pixel-level energy minimization methods, capturing low-level appearance features through unary potentials and ensuring consistent segmentation results with pairwise ones~\cite{wang2023segrefiner}. 
In recent years, many studies leverage prompts as input features and inject them into the models to produce segmentation results~\cite{sam,sam2,Semantic-SAM,SEEM}. The Segment Anything Model (SAM)~\cite{sam} pre-trained on a large-scale datasets has emerged as a benchmark for interactive segmentation. 
Considering that SAM cannot perform semantic prediction, many works~\cite{Semantic-SAM,SEEM} utilize category labels to further fine-tune SAM and enable it to achieve the semantic segmentation task. For example, SEEM~\cite{SEEM} further trains SAM using labeled segmentation data through a bipartite matching constraint, thus providing the model with the ability to predict semantics. Semantic-SAM~\cite{Semantic-SAM} proposes a multi-choice learning scheme via multi-task training across diverse datasets, enabling the model to segment at various granularities while predicting semantic labels.
Other works~\cite{ma2024segment,xu2024embodiedsam,Gaussian-Grouping} apply SAM to diverse specific domains to solve application tasks. MedSAM~\cite{ma2024segment} fine-tunes SAM by collecting a large-scale dataset of medical image samples in multiple imaging modalities and types of cancer, achieving superior results in disease segmentation.
Gaussian-Grouping~\cite{Gaussian-Grouping} leverages the SAM as supervision model to jointly train a reconstruction and segmentation of objects in open-world 3D scenes for high-quality 3D edit.
% For instance, SAM functions as a robust pretrained encoder-decoder model that can be fine-tuned for specific tasks~\cite{ravi2024sam,yang2024samurai}. Some other approaches maintain the interactive nature of segmentation by focusing on parameter-efficient fine-tuning techniques~\cite{xu2024embodiedsam}. 
% To achieve robust segmentation across diverse scenarios, some methods fine-tune SAM’s decoder using large-scale domain datasets~\cite{houlsby2019parameter,ma2024segment}. 
% Although these approaches show notable advantages across various benchmarks, they have yet to be consistently evaluated against similar methods, and a comprehensive analysis of how different interaction strategies influence outcomes is still lacking.

%-------------------------------------------------------------------------
% %%%%%%%%%%%% figure 3
\begin{figure*}
\begin{center}
\setlength{\belowcaptionskip}{-0.1cm}
\includegraphics[width=0.98\linewidth]{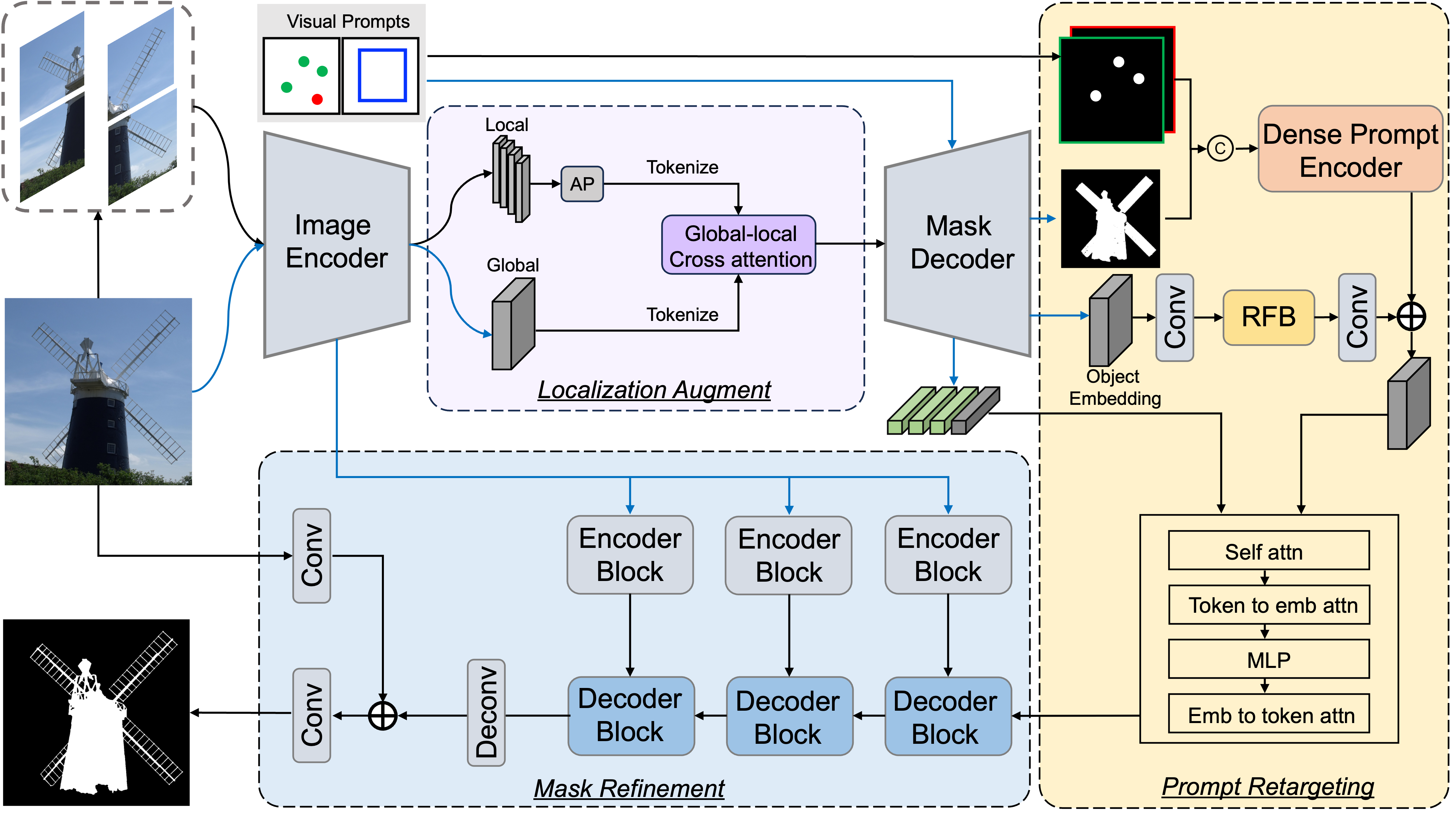}
\caption{
An overview of the proposed framework SAM2Refiner. It contains a localization augment module to balance the detailed and semantic representations, a prompt retargeting for enhancing response of input prompts and a mask refinement structure to boost the quality of mask outcomes. The blue line denotes the SAM2 pipeline, and the black line denotes our SAM2Refiner pipeline.
}
\label{fig:network}
\end{center}
\end{figure*}
%-------------------------------------------------------------------------

\subsection{High-Quality Segmentation}
High-quality segmentation as a fundamental computer vision task focuses on accurately segmenting various complex and detail-rich objects, which ranges from many sub-tasks including semantic segmentation~\cite{long2015fully,zhao2017pyramid}, instance segmentation~\cite{he2017mask,dai2016instance} and panoptic segmentation~\cite{kirillov2019panoptic}. 
% Recent researches like SegGPT~\cite{SegGPT} enable diverse segmentation tasks through in-context visual learning.
% Existing methods for high-quality segmentation are tailored to specific tasks, such as image and video semantic segmentation~\cite{lin2017refinenet,zhao2018icnet,takikawa2019gated,yuan2020segfix}, instance segmentation~\cite{transfiner,vmt,tang2021look} or panoptic segmentation~\cite{de2023intra}. 
Traditional methods~\cite{Panoptic-deeplab,deeplab-v3,hrnet,medseg1,cascadepsp} on high-quality segmentation dedicate to designing elaborate structures upon on CNN-based networks in order to capture both low-level local patterns and semantic global features through various reception fields towards holistic representation.
Deeplab-v3~\cite{deeplab-v3} designs a set of multi-level convolution kernels to capture features with diverse reception fields and HRNet~\cite{hrnet} employs cascaded structures to fuse features from different layers to learn both low-level and high-level information.
% More recently, many approaches focus on post-segmentation refinement using graphical models like CRF~\cite{krahenbuhl2011efficient} and other utilize separate deep networks for iterative improvements~\cite{cascadepsp,shen2022high}.
More recently, many approaches~\cite{swin-unet,SegFormer,SegGPT,maskformer,mask2former} built on the transformer backbone focus on introducing local patterns into the model by narrow the reception field of self-attention module. For example, Mask2Former~\cite{mask2former} presents an efficient versatile architecture by using masked attention to extract local features for high-quality image segmentation tasks. Swin-Unet~\cite{swin-unet} adopts the Swin-Transformer network to the UNet like structures to adaptively capture object features with various scales via hierarchical attention.
Latest, the Pi-SAM~\cite{liu2024segment} designs a mask decoder to expands SAM’s ability at high resolution images, however, it may diminish the response of the input prompt and exhibit biased preferences for background clicks. 
% In contrast, our work emphasizes accurately segmenting diverse objects in new data using flexible prompting. We propose HQ-SAM, a high-quality zero-shot segmentation model that generalizes across various tasks and domains. Unlike post-segmentation refinement approaches, HQ-SAM directly predicts high-quality masks by reusing the image encoder and mask decoder from SAM, rather than relying on a coarse mask as input for a separate refinement network. Our model architecture builds upon SAM with minimal overhead, incorporating efficient token learning for accurate mask predictions. We demonstrate its effectiveness through extensive zero-shot experiments.

\section{Method}
\label{sec:method}

We propose SAM2Refiner, which effectively generates fine-grained segmentation masks with strong perception for both global semantic and local details,
and simultaneously maintaining stable prompting ability. In Sec~\ref{sec:3.1}, we first briefly review the architecture of SAM and SAM2.
Then, in Sec~\ref{sec:3.2}, we introduce the three modules proposed upon SAM2's backbone to effectively improve the precision for segmentation.
Finally, in Sec~\ref{sec:3.3}, we describe the training and inference process of our method, we also depict how our method can be seamlessly inserted into SAM2's video streaming pipeline for zero-shot inference. The overall framework of our SAM2Refiner is shown in Figure~\ref{fig:network}.

\subsection{Preliminaries: SAMs}
\label{sec:3.1}

SAM consists of an image encoder, a prompt encoder, and a mask decoder. The image encoder works
for image feature extraction with a ViT~\cite{alexey2020image} backbone. The prompt encoder
is employed to encode the points, boxes, or masks into prompt tokens. The mask decoder utilizes a
two-layer transformer-based decoder that takes both the extracted image embedding with the concatenated
output and prompt tokens for final mask prediction.

SAM2 extends SAM’s ability to support both images and videos. For image setting, the three modules are largely
follow SAM except for two differences, firstly, SAM2 adopt an MAE pretrained Hiera~\cite{ryali2023hiera} image
encoder, which is hierarchical compared with the plane structure in SAM's encoder, and secondly, to solve ambiguity in
images without valid object, SAM2 predicts an additional token in mask decoder to indicate whether the object of interest is present.
SAM2 introduces additional memory modules to support video streaming, including memory attention, memory encoder, 
and memory bank. The role of memory attention is to condition the current frame features on the past frame
features and predictions as well as on any new prompts. The memory encoder is employed to obtain a memory embedding from
masks generated from mask decoder, these memory embeddings are appended to a memory bank in a first-in-first-out (FIFO) way.

\subsection{Ours: SAM2Refiner}
\label{sec:3.2}
In this section, we describe the architecture of SAM2Refiner, which is built upon SAM2's backbone.
This design offers two key advantages, 1) the hierarchical image encoder facilitates multi-scale feature capturing, which aids in feature refinement.
2) it allows for flexible integration into the video streaming pipeline.
The proposed architecture comprises three modules, Localization Augment (LA) is inserted into the middle layer of image encoder and mask decoder to refine
the details of global features with local contextual feature
cues. Prompt Retargeting (PR) aligns the prompt feature with enhanced object embedding to provide flexible prompting. Mask Refinement (MR) further incorporates object features and multi-scale features to generate higher-resolution masks with finer details.

\subsubsection{Localization Augment}

We evenly crop the input image $\mathbf{I} \in \mathbb{R}^{B \times 3 \times H \times W}$ into four non-overlapping
sub-images $\{\mathbf{s}_m\}_{m=1}^4 \in \mathbb{R}^{B \times 3 \times \frac{H}{2} \times \frac{W}{2}}$. To align
with input resolution of SAM2's image encoder, we upsample the sub-images to the same resolution with $\mathbf{I}$ and obtain
$\{\mathbf{S}_m\}_{m=1}^4 \in \mathbb{R}^{B \times 3 \times H \times W}$. $\mathbf{I}$ and $\{\mathbf{S}_m\}_{m=1}^M$ are then fed in batches to
the image encoder and generate the highest-level feature maps $\{{F}_i\}_{i=1}^5 \in \mathbb{R}^{B \times C \times \frac{H}{16} \times \frac{W}{16}}$.
We then split the feature maps into global feature $G = F_1$ and local feature $\{L_{m}\}_{m=1}^4 = \{F_i\}_{i=2}^5$.
The local features contain fine-grained details of input images and are used to enhance global feature with detailed clues. The local features are first assembled into a unified concatenated feature $L \in \mathbb{R}^{B \times C \times \frac{H}{8} \times \frac{W}{8}}$
following their respective positions in the original image. We then apply a multi-granularity average pooling operation as ~\cite{yu2024multi}
to capture important contextual feature cues with different scales. The pooling functions are denote as:
\begin{equation}
  \begin{aligned}
    P_i & = \mathtt{AvgPool}_i(L), i \in \{1,...,n\}. \\
  \end{aligned}
\end{equation}
Where $\mathtt{AvgPool}_i$ is the $i$-th pooling function, $P_i$ is the $i$-th feature map. We set $n=3$ in our experiments,
and the receptive fields for each pooling function are set 2, 4, 8 respectively. The feature maps $P_i$ are then tokenized and concatenated into a multi-granularity local features $L_t$. For global feature $G$, we tokenized it as $G_t$. We conduct global-local cross attention between $G_t$ and $L_t$ to obtain the refined global feature $G_{re}$:
\begin{equation}
  \begin{aligned}
      G_{ref}& = \mathtt{CrossAttn}(G_t, L_t, L_t) \\
  \end{aligned}
\end{equation}
Before sending into attention layer, the positional encoding is added to the features to access critical geometric information. $G_{re}$ is able to highlight the local information about the object within the global representation.

%-------------------------------------------------------------------------
\begin{figure*}
\begin{center}
\includegraphics[width=0.98\linewidth]{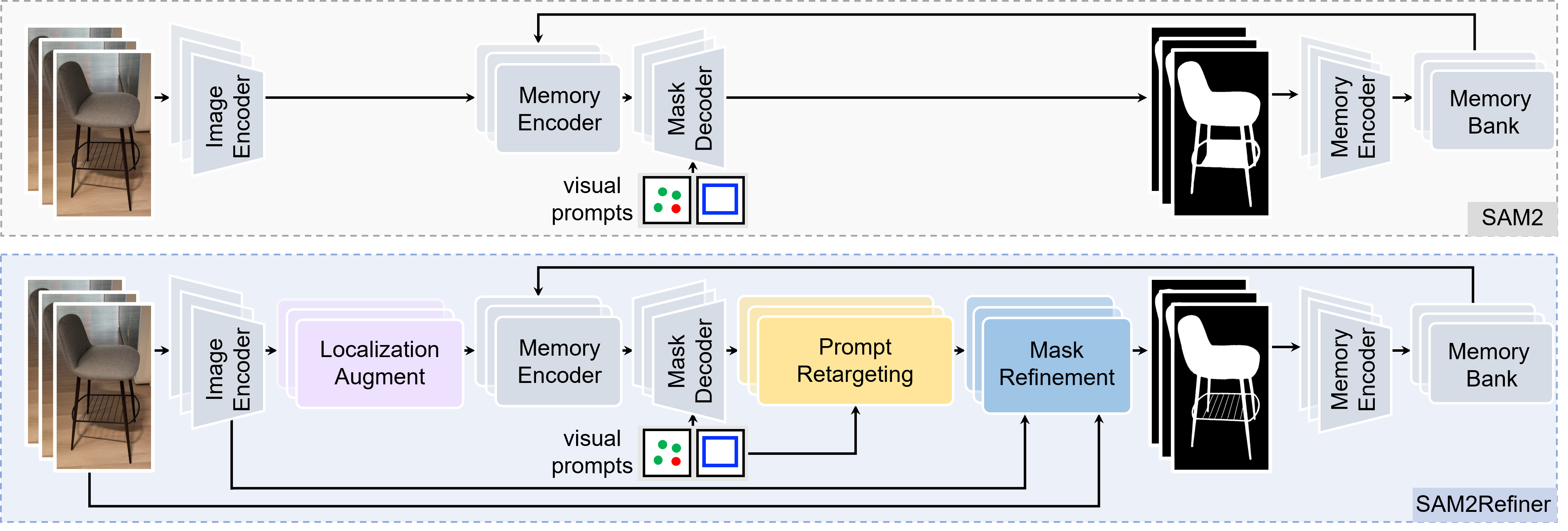}
\caption{A Comparison between SAM2 and SAM2Refiner. SAM2Refiner introduces three modules for fine-grained interactive segmentation, which can be seamlessly embedded into SAM2's video streaming pipeline to support both images and videos.}
\label{fig:video-seg}
\end{center}
\end{figure*}
%-------------------------------------------------------------------------

\subsubsection{Prompt Retargeting}

We replace the global feature $G$ with $G_{re}$ as the input to the mask decoder. The output mask $M \in \mathbb{R}^{B \times 3 \times h \times w}$
and object embedding $E \in \mathbb{R}^{B \times C \times h \times w}$ of the mask decoder therefore contain richer representations
of the local details. To enhance the discriminability of features in small patches,
we introduce a modified version of multi-branch convolutional block RFB~\cite{liu2018receptive}, which applies different kernels to achieve multi-size receptive fields. To be specific, we first employ the bottleneck structure in each branch, consisting of a $1 \times 1$ convolutional layer to decrease the channel size in object embedding. Secondly, we apply three parallelized blocks with $1 \times n$ plus a $n \times 1$ convolutional layer to each branch,
where we set $n = 3,5,7$, respectively. Thirdly, three branches are concatenated and encoded via a $1 \times 1$ convolutional layer.
Subsequently, we obtain the augmented embedding $E_a$ by adding the concatenated embedding and a shortcut embedding encoded from another $1 \times 1$ convolutional layer.
Although the ability to perceive local details is significantly enhanced, $E_a$ exhibits reduced sensitivity to the input visual prompts.
To maintain a stable prompting ability towards the enhanced embedding, we employ a three-channel prompt map to align the prompt feature with the augmented object embedding $E_a$.
The map resolution is set as $256 \times 256$ to reduce the computational cost and only two visual prompts (clicks and masks) are encoded.
We represent clicks as a fixed size radius, where positive clicks are encoded in the first channel, and negative clicks are encoded in the second channel. The mask $M$ is then concatenated into the third channel to form a three-channel dense prompt map $D_m$. We propose a dense prompt encoder to transform the $D_p$ to a dense prompt embedding $E_p$. The dense prompt encoder consists of two $2 \times 2$ convolution layers followed by GELU activations and layer normalization, and one $1 \times 1$ convolution layer to map the channel dimension to 256. The dense prompt embedding $E_p$ is added to $E_a$ and obtain $E_{ap}$ in an element-wise manner, followed by a retargeting modules with several attention operations as in SAM2's mask decoder.
We re-use the sparse prompt embedding and output token from SAM2's mask decoder to provide positional and class information for the renewing of the object embedding. The retargeting module contains four steps, (1) self-attention on the tokens concatenated from the the sparse prompt embedding and output token; (2) cross-attention from token to $E_{ap}$; (3) MLP updated on token, (4) cross-attention from $E_{ap}$ to token. The renewed object embedding allows to possess a higher sensitivity towards the input visual prompts.

%%%%%%%%%%%
%-------------------------------------------------------------------------
\begin{table*}[t]
\centering
\caption{Results comparison with state-of-the-art methods on four fine-grained image segmentation tasks.}
% We adopt the boxes converted from their GT masks as the box prompt input.
\vspace{0.05in}
\resizebox{1.0\linewidth}{!}{
\begin{tabular}{l|cc|cc|cc|cc|cc}
\toprule
   \multirow{2}{*}{Model} & \multicolumn{2}{c|}{DIS}& \multicolumn{2}{c|}{COIFT}& \multicolumn{2}{c|}{HRSOD}& \multicolumn{2}{c|}{ThinObject} & \multicolumn{2}{c} {Average}  \\
   \cmidrule(lr){2-3}\cmidrule(lr){4-5}\cmidrule(lr){6-7}\cmidrule(lr){8-9}\cmidrule(lr){10-11}
   % \cline{2-9}
& mIoU & mBIoU & mIoU & mBIoU & mIoU & mBIoU   & mIoU & mBIoU  & mIoU & mBIoU \\
\midrule
SAM (vit\_l)~\cite{sam} & 55.5 & 47.9 & 88.4 & 83.4 & 85.2 & 78.4 & 66.1 & 55.6 & 73.8 &  66.3 \\
SAM (vit\_h)~\cite{sam} & 50.6 & 43.9 & 86.5 & 81.7 & 80.4 & 74.2 & 57.6 & 49.4 & 68.8 & 62.3  \\
SAM2 (hiera\_b+)~\cite{sam2} & 57.2 & 49.9 & 93.2 & 88.7 & 85.2 & 76.3 & 66.3  & 65.1  & 75.5 & 70.0\\
SAM2 (hiera\_l)~\cite{sam2} & 57.6 & 50.7 & 92.4 & 88.0 & 86.3 & 79.5 & 77.8  & 67.7  & 78.5 & 71.5\\
HQ-SAM (vit\_l)~\cite{hq-sam} & 77.4 & 69.2 & 94.4 & 89.7 & 91.9 & 84.6 & 87.0 & 76.6 & 87.7 & 80.0 \\
HQ-SAM (vit\_h)~\cite{hq-sam} & 77.9 & 69.7 & 94.4 & 89.5 & 91.9 & 83.5 & 88.5 & 78.6 & 88.2 & 80.3 \\
\midrule
SAM2Refiner (hiera\_b+) & 83.4 & 78.5 & 96.5 & 93.3 & 94.1 & 89.4 & 95.1 & 90.1 & 92.3 & 87.8 \\
SAM2Refiner (hiera\_l)  & \textbf{85.2} & \textbf{80.6} & \textbf{96.6} & \textbf{93.6} & \textbf{94.7}& \textbf{90.2} & \textbf{95.9} & \textbf{91.0} & \textbf{93.1} & \textbf{88.9} \\

 \bottomrule
\end{tabular}}
\vspace{-0.1in}
\label{tab:quantitative_cmp}
\end{table*}
%-------------------------------------------------------------------------

\subsubsection{Mask Refinement}

The final mask of SAM2 is directly interpolated from the output prediction of its mask decoder, with a resolution of $256 \times 256$.
This interpolation process may inevitably damage the details of the output prediction, particularly in high-resolution images.
Our mask refinement module is designed to generate higher-resolution masks while enhancing the result quality. We take the object embedding from the prompt retargeting module and the multi-scale features from SAM2's hierarchical features to predict the final mask.
The encoder blocks in Figure 3 represent the sub-modules within the image encoder, which generate multi-scale features with resolutions $256 \times 256$, $128 \times 128$ and $64 \times 64$, respectively. We incorporate the UNet-like structure to fuse the object features and multi-scale image features. Each of decoder blocks contains an upsampling layer and a convolutional layer, and the upsampling feature along with image feature are concatenated via the channel dimension before sending into the convolutional layer. After the final decoder block, we conduct two transposed convolution layers to obtain mask feature with $1024 \times 1024$ resolution. The low-level feature generated from input image $\mathbf{I}$ is added further enhance image quality, and a convolutional layer is then used to generate the final segmentation map.

\subsection{Training and Inference}
\label{sec:3.3}
\noindent {\bf Training.} We train our SAM2Refiner on image datasets only. We utilize the HQSeg-44K~\cite{hq-sam} as the training database, which includes six image datasets with extremely fine-grained mask labeling, including DIS~\cite{qin2022}(train set), ThinObject-5K~\cite{liew2021deep} (train set), FSS-1000~\cite{li2020fss}, ECSSD~\cite{shi2015hierarchical}, MSRA-10K~\cite{cheng2014global}, DUT-OMRON~\cite{yang2013saliency}. During training, we fix the model parameters of the pre-trained SAM2 model. We train by sampling mixed types of prompt including bounding boxes, randomly sampled points (either positive or negative), and coarse mask input. We use a learning rate of 0.001 and train for 20 epochs, with a learning rate decay schedule. We train on 8 NVIDIA V100 GPUs with a total batch size of 32. If not otherwise specified, we use the hierarchical large model as the default backbone.

\noindent {\bf Loss Function.} Following HQ-SAM~\cite{hq-sam}, we use a combination of binary cross-entropy (BCE) loss and dice loss for supervision.
In addition to calculating the loss at the final prediction, we introduce intermediate supervision at each layer of the three decoder blocks.
Before the loss calculation, each of the output features is sent to a separate convolutional layer to align the channel size with ground-truth mask.
Our final loss function is:
\begin{equation}
  \begin{aligned}
      \mathcal{L} = \lambda_{f}(\mathcal{L}_{f}^{bce} + \mathcal{L}_{f}^{dice}) + \lambda_{i}\sum_{i=1}^{3}(\mathcal{L}_{i}^{bce} + \mathcal{L}_{i}^{dice}) \\
  \end{aligned}
\end{equation}
where $\lambda_{f}$ and $\lambda_{i}$ are set to 1.0 and 0.3 in our experiment.

\noindent {\bf Inference.} 
For image setting, we follow the same inference pipeline as SAM. In the SAM2 temporal streaming pipeline, we extend our SAM2Refiner to the video segmentation by adding the cross-frame interaction via the memory bank. As suggested in Figure~\ref{fig:video-seg}, compared to the original SAM2 pipeline, SAM2Refiner, including three proposed components, leverages the mask feature memory of previous frames to perform segmentation for the current frame, achieving accurate video segmentation. Note that for each frame except the first frame, the model re-uses the prompt input from the first frame as the original prompt to perform frame-by-frame segmentation.

%-------------------------------------------------------------------------

\section{Experiments}
\label{sec:exp}

\subsection{Experimental Setup}
% \parsection{Training Datasets} detailed the datasets composition using a small table.

\noindent\textbf{Datasets.} 
Following the prior work~\cite{hq-sam}, we evaluate the model ability in both image and video segmentation tasks respectively. In image segmentation, all models are assessed in four challenging fine-grained segmentation benchmarks, including DIS-validation~\cite{qin2022}, ThinObject5K-test~\cite{liew2021deep}, COIFT~\cite{liew2021deep} and HR-SOD~\cite{zeng2019towards}. For video object segmentation, we leverage two long-term video segmentation datasets, LVOS-v1 and LVOS-v2~\cite{gupta2019lvis}, to evaluate our framework.

\noindent\textbf{Evaluation Metrics.} 
To evaluate the results of the image segmentation task, we echo previous works~\cite{hq-sam,sam} and adopt mIoU and mBIoU scores as metrics. For video object segmentation, we utilize $J$ (region similarity), $F$ (contour accuracy), and the combined $J\&F$ scores as metrics, where evaluations are performed in
a semi-supervised setting with the first frame prompt given.

\subsection{Zero-Shot Comparison}
We perform extensive zero-shot transfer comparisons between our SAM2Refiner, HQ-SAM and SAM with various backbones on 6 benchmarks across image and video segmentation tasks. We do not compare Pi-SAM as they have not open-sourced their code.
% including image segmentation datasets DIS (val)~\cite{qin2022}, COIFT~\cite{liew2021deep}, HRSOD~\cite{zeng2019towards}, ThinObject-5k (test)~\cite{liew2021deep} and video segmentation datasets LVOS-v1 and LVOS-v2~\cite{gupta2019lvis}.

\begin{figure*}
\begin{center}
\includegraphics[width=0.98\linewidth]{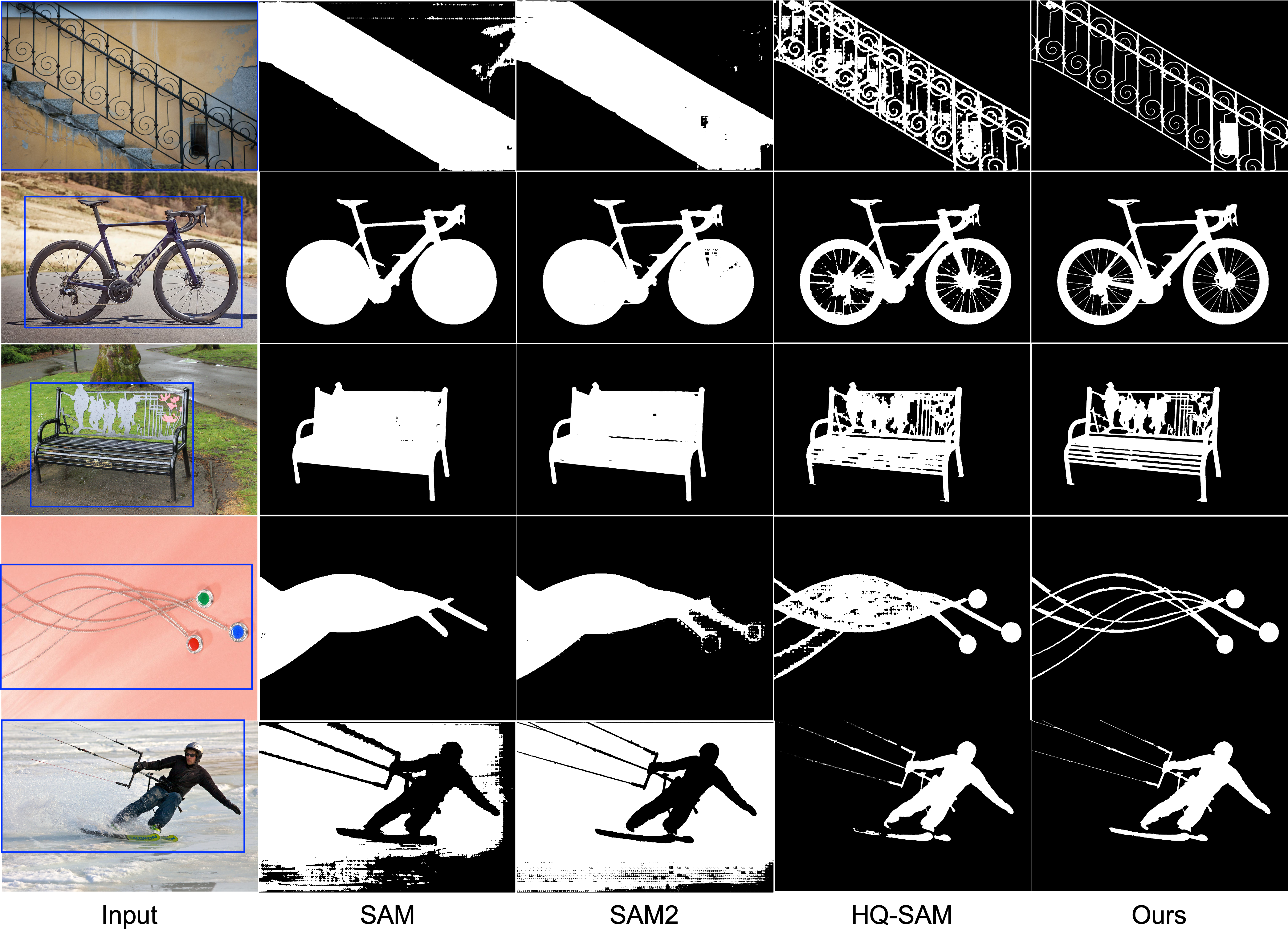}
\caption{Qualitative comparison with previous methods. Given the blue box as visual prompt, our proposed SAM2Refiner produces more accurate results with correct structure and clear boundaries. Zoom in for better visualization.}
\label{fig:qualitative}
\end{center}
\end{figure*}

% \noindent\textbf{Results on the SGinW Benchmark} Equipped with the same Grounding-DINO~\cite{liu2023grounding} as box prompts, we also performed experiments by replacing SAM with HQ-SAM in Grounded-SAM, and obtained \textbf{the first place} in the Segmentation in the Wild (SGinW) competition\textcolor{red}{$^1$} on the zero-shot track. Note that SGinW contains \textit{\textbf{25} zero-shot in-the-wild segmentation datasets} for evaluation, and Grounded-HQ-SAM with 49.6 mean AP and outperforms Grounded-SAM obviously using the same detector. 

\noindent\textbf{Zero-Shot Open-world Image Segmentation.} 
To evaluate the zero-shot segmentation results in the open-world environment, we compare our SAM2Refiner with previous methods on four image segmentation benchmarks. 
In Table~\ref{table2-2}, our SAM2Refiner with large backbone achieves state-of-the-art performance of 85.2\%, 96.6\%, 94.7\%, 95.9\% mIoU scores and 80.6\%, 93.6\%, 90.2\%, 91.0\% mBIoU scores on DIS, COIFT, HRSOD and ThinObject datasets, respectively. In average, SAM2Refiner with a smaller backbone (hiera$\_$l) performs 93.1\% and 88.9\% average mIoU and mBIoU scores, significantly outperforming the second-best one HQ-SAM (vit$\_$h) with 4.9\% and 8.6\% scores.
To clear represent the segmentation results, we show the comparison of the visualization results of segmentation maps. From Figure~\ref{fig:qualitative}, our SAM2Refiner can produce much more accurate segmentation maps than other comparisons, specifically on detailed objects such as boundary and small regions.
These results demonstrate that the SAM2Refiner benefitting from the proposed modules achieves superior fine-grained image segmentation and strong generalization ability.

%-------------------------------------------------------------------------
\begin{table}[t]
\renewcommand\arraystretch{1}
\setlength{\tabcolsep}{2.2pt}
\centering
% \scriptsize
\footnotesize
\caption{Comparison of the proposed method with prior works on video object segmentation task.}
% \vspace{2pt}
\label{table2-2}
\begin{tabular}{cccc|ccccc}
\toprule
 \multicolumn{1}{c}{\multirow{1}{*}{Methods}} & \multicolumn{3}{c|}{LVOS v1} & \multicolumn{5}{c}{LVOS v2}  \\  % \cmidrule{5-10}
 &  $\mathcal{J}\&\mathcal{F}$ & $\mathcal{J}$  & $\mathcal{F}$  &  $\mathcal{J}\&\mathcal{F}$ & $\mathcal{J}_s$  & $\mathcal{F}_s$ & $\mathcal{J}_u$ & $\mathcal{F}_u$  \\ \midrule
LWL~\cite{STCN}                     & 56.4 & 51.8 &   60.9    & 60.6 & 58.0 & 64.3 & 57.2 & 62.9  \\ % \midrule
CFBI~\cite{STCN}                    & 51.5 & 46.2 &   56.7    & 55.0 & 52.9 &  59.2  & 51.7 & 56.2  \\ 
STCN~\cite{STCN}                    & 48.9 & 43.9 &   54.0    & 60.6 & 57.2 &  64.0  & 57.5 & 63.8  \\ 
RDE~\cite{RDE}                      & 53.7 & 48.3 &   59.2    & 62.2 & 56.7 &  64.1  & 60.8 & 67.2  \\ 
SwinB-DeAOT-L~\cite{SwinB-DeAOT-L}  & - & - &   -    & 63.9 & 61.5 &  69.0  & 58.4 & 66.6  \\  
XMem~\cite{XMem}                    & 52.9 & 48.1 &   57.7    & 64.5 & 62.6 &  69.1  & 60.6 & 65.6  \\   \midrule
SAM2~\cite{sam2}                    & 77.2 & 72.2 &   82.2    & 78.3 & 75.2 &  87.0 & 68.5 &  79.4 \\
HQ-SAM2~\cite{hq-sam}               & 79.6 & {74.9} & 84.3  & 79.1 & 76.5 & \textbf{88.3} &  69.4 &  \textbf{82.2}  \\
% SAM2Refiner                         & \textbf{80.3} & 74.4 &   \textbf{86.3}    & \textbf{79.6} & \textbf{78.5} & 87.7 &  \textbf{71.3} &  81.1  \\  
SAM2Refiner                         & \textbf{80.9} & \textbf{75.1} & \textbf{86.8} & \textbf{79.9} & \textbf{78.7} & 87.9 &  \textbf{71.6} &  81.5  \\  
\bottomrule
\end{tabular}
\end{table}
%-------------------------------------------------------------------------

% \noindent\textbf{Zero-Shot Segmentation on High-resolution BIG Dataset} In Table~\ref{table2-2}, we compare the zero-shot segmentation quality between SAM and SAM2Refiner on the high-resolution BIG benchmark~\cite{cascadepsp} with two types of prompts, including using GT object boxes or the provided coarse masks input. SAM2Refiner consistently surpasses SAM, with obvious advantages using different types of prompts, and is much more robust to coarse masks prompts with partial boundary errors (provided by PSPNet~\cite{zhao2017pyramid}).

\noindent\textbf{Zero-Shot Open-World Video Object Segmentation.} To evaluate the ability of the proposed method on video segmentation, we extend our model on semi-supervised video object segmentation task upon the pipeline of SAM2. Towards the zero-shot segmentation in the open-world scenario, we compare our model with previous video segmentation works in two benchmarks, including LVOS v1 and LVOS v2. As suggested in Table~\ref{table2-2}, our model yields 80.9\%, 75.1\%, 86.8\% scores of $\mathcal{J}\&\mathcal{F}$, $\mathcal{J}$, $\mathcal{F}$ and 79.9\%, 78.7\%, 87.9\%, 71.6\%, 81.5\% of $\mathcal{J}\&\mathcal{F}$, $\mathcal{J}_s$, $\mathcal{F}_s$, $\mathcal{J}_u$, $\mathcal{F}_u$ in two datasets respectively, significantly outperforming over previous comparisons~\cite{STCN,RDE,SwinB-DeAOT-L,XMem,sam2,hq-sam}. These results demonstrate that the proposed framework can not only enhance fine-grained image segmentation, but also boost the video segmentation with refined object details.

% \subsection{Detailed Analysis}
\noindent\textbf{Analysis on prompt number.}
To evaluate the response intensity to input prompts in SAM2Refiner compared to other models, we input varying numbers of point prompts into the model in image segmentation task. In Table~\ref{tab:analysis-point-num}, we observe that the average segmentation results decline as the number of prompt points decreases. However, our SAM2Refiner indicates consistent advantages in segmentation accuracy across varying numbers of point prompts, while SAM, SAM2 and HQ-SAM reflect relatively low results. This clearly suggests that the proposed method with the capability of prompt re-targetting can effectively achieve a strong response for input prompts.

%-------------------------------------------------------------------------
\begin{table}[t]
\centering
\caption{Analysis of the point prompt response in interactive image segmentation.}
\vspace{-0.05in}
\resizebox{1.0\linewidth}{!}{
\begin{tabular}{cc|cccc}
\toprule
   \multirow{1}{*}{Point Num.} & \multicolumn{1}{c|}{Metrics}& \multicolumn{1}{c}{SAM}& \multicolumn{1}{c}{SAM2}& \multicolumn{1}{c}{HQ-SAM} & \multicolumn{1}{c} {SAM2Refiner}  \\ \cline{1-6}
\multirow{2}{*}{1} & mIoU    & 8.7  & 11.4 & 17.7  & \textbf{25.2} \\
& mBIoU                      & 6.3  & 9.9  & 14.3  & \textbf{20.6} \\ \midrule
\multirow{2}{*}{2} & mIoU    & 30.8 & 35.6 & 44.6  & \textbf{52.5} \\
& mBIoU                      & 26.2 & 31.7 & 40.2  & \textbf{47.4} \\ \midrule
\multirow{2}{*}{5} & mIoU    & 70.1 & 73.9 & 84.2  & \textbf{90.0} \\
& mBIoU                      & 62.3 & 66.4 & 75.5  & \textbf{84.9} \\ \midrule
\multirow{2}{*}{10} & mIoU   & 73.8 & 78.5 & 87.7  & \textbf{93.1} \\
& mBIoU                      & 66.3 & 71.5 & 80.0  & \textbf{88.9} \\
\bottomrule
\end{tabular}}
\label{tab:analysis-point-num}
\end{table}

\begin{table}[t]
\centering
\caption{Ablation study on the three modules of SAM2Refiner, including Localization Augment~(LA), Prompt Retargeting~(PR) and Mask Refinement~(MR). The results are averaged over the four evaluated datasets.}
\vspace{-0.05in}
\resizebox{0.9\linewidth}{!}{
\begin{tabular}{l|c|c|c|cc}
\toprule
   & LA & PR & MR  & mIoU & mBIoU  \\ \midrule
SAM2~\cite{ravi2024sam} &  &    &  & 78.5 & 71.5 \\ \midrule
 & $\checkmark$ &  &    & 85.0 & 67.4  \\
&   & $\checkmark$     & $\checkmark$ & 91.1 & 86.0   \\
SAM2Refiner &  $\checkmark$ &    &   $\checkmark$ & 92.7  & 88.3 \\
 &  $\checkmark$ & $\checkmark$  &    &  91.4 & 86.2\\
 & $\checkmark$ & $\checkmark$   & $\checkmark$ &  \textbf{93.1} & \textbf{88.9} \\ \bottomrule
\end{tabular}}
\vspace{-0.2in}
\label{tab:ablation}
\end{table}

\subsection{Ablation Study}
We conduct ablation experiments to evaluate the effectiveness of the proposed modules within our SAM2Refiner, i.e., Localization Augment (LA), Prompt Retargeting (PR) and Mask Refinement (MR), and we study the affect of the number of input prompts. 

\begin{figure}
\begin{center}
\includegraphics[width=0.98\linewidth]{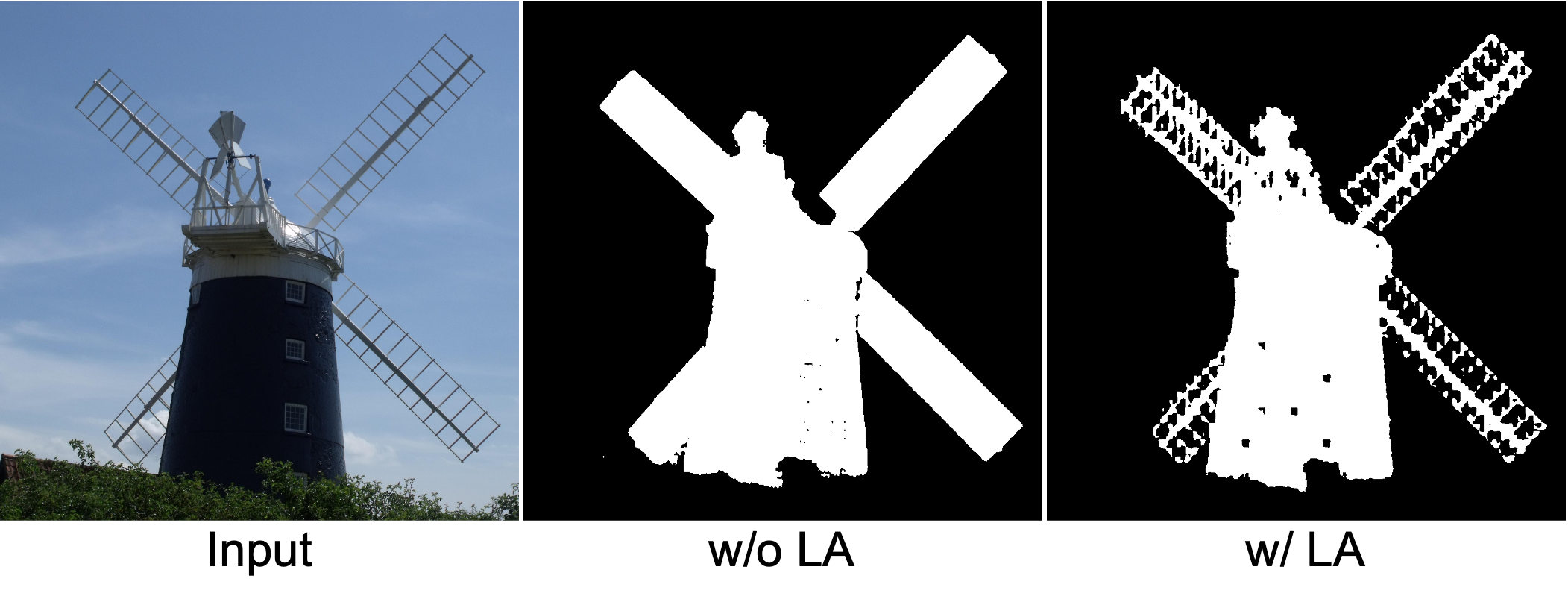}
    \vspace{-10pt}
\caption{Effectiveness of LA module. LA suggest remarkable improvement on local details although it displays some flickering boundaries.}
\label{fig:ab1}
\end{center}
    \vspace{-5pt}
\end{figure}

\begin{figure}
\begin{center}
    \vspace{-10pt}
\includegraphics[width=0.98\linewidth]{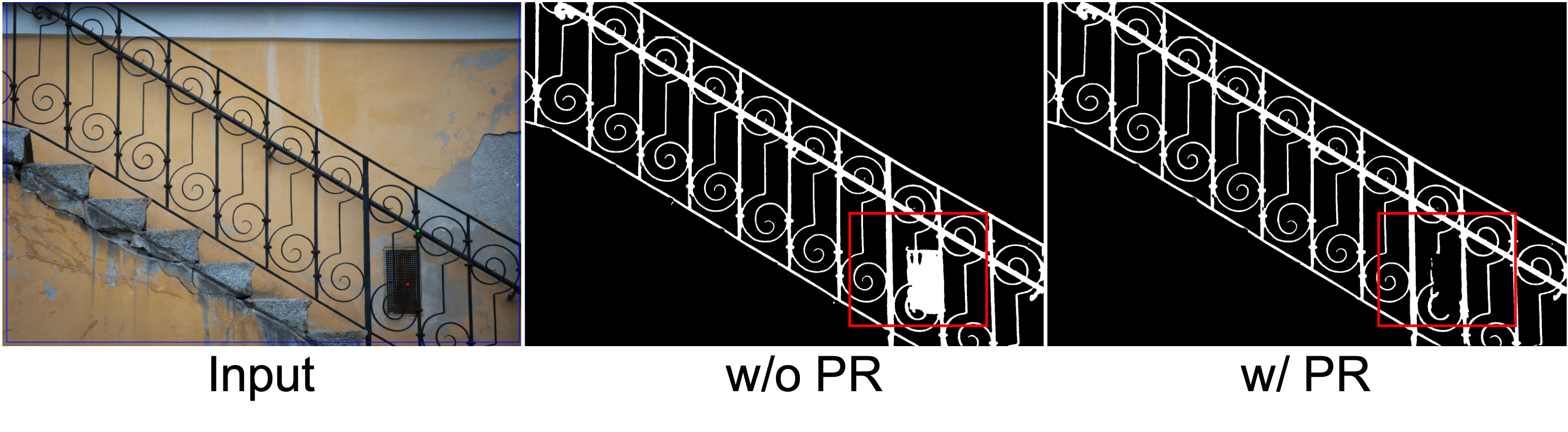}
    \vspace{-10pt}
\caption{Effectiveness of PR module. PR allows precise response for point prompts in ambiguous regions. Green point and red point denote positive and negative prompt, respectively.}
\label{fig:ab2}
\end{center}
\end{figure}

\noindent\textbf{Effectiveness of the proposed components.} 
In the LA module, the original global features attend to local features from sub-images, thereby enriching the representation with more detailed local information. As shown in Table~\ref{tab:ablation}, compared to SAM2, the LA module markedly enhances the mIoU from 78.5\% to 85.0\%, despite a slight decrease in boundary accuracy. This is because the LA module strengthens the perception of fine structures, but this can come at the cost of boundary precision. Figure~\ref{fig:ab1} verifies this phenomenon, showing that results with LA exhibit more detailed local features but also show flickering boundaries. The second to fourth rows of the SAM2Refiner in Table~\ref{tab:ablation} highlight the effectiveness of each individual module for fine-grained segmentation. The PR module allows the model to maintain strong sensibility towards point prompting in intricate regions. As shown in Figure~\ref{fig:ab2}, result with PR module is able to correctly respond to the input visual prompt. Compared to PR, both the LA and MR modules deliver greater performance improvements in mIoU and mBIoU, with about 2 points gain in mIoU and 3 points gain in mBIoU, underscoring their crucial roles in enhancing segmentation precision. 

% For ablation experiments, we use the four aforementioned extremely accurate segmentation datasets, namely, DIS (val)~\cite{qin2022}, ThinObject-5K (test)~\cite{liew2021deep}, COIFT~\cite{liew2021deep} and HR-SOD~\cite{zeng2019towards} as well as the COCO validation set.

% \noindent\textbf{Effectiveness of the Localization Augment}

% HQ-SAM employs HQ-Output Token for high-quality mask prediction. Table~\ref{table2-2} compares our HQ-Output Token to the baseline SAM and other existing prompt/token learning strategies, such as adding an additional three context tokens~\cite{zhou2022coop} as learnable vectors into the SAM's mask decoder for better context learning. 

% \noindent\textbf{Influence of the Prompt Retargeting}
% HQ-SAM employs HQ-Output Token for high-quality mask prediction. Table~\ref{table2-2} compares our HQ-Output Token to the baseline SAM and other existing prompt/token learning strategies, such as adding an additional three context tokens~\cite{zhou2022coop} as learnable vectors into the SAM's mask decoder for better context learning. 

% \noindent\textbf{Affect of the mask refinement}
% HQ-SAM employs HQ-Output Token for high-quality mask prediction. Table~\ref{table2-2} compares our HQ-Output Token to the baseline SAM and other existing prompt/token learning strategies, such as adding an additional three context tokens~\cite{zhou2022coop} as learnable vectors into the SAM's mask decoder for better context learning.

%-------------------------------------------------------------------------

\section{Conclusion}
\label{sec:conclusion}
In this work, we propose SAM2Refiner to enhance interactive segmentation for images and videos. SAM2Refiner balances local detail perception and prompting stability, achieving fine-grained visual representation. By integrating local contextual cues through a cross-attention mechanism and employing a UNet-like structure for feature fusion, SAM2Refiner generates high-quality segmentation masks. In addition, a prompt retargeting module is designed to improve the alignment of object embeddings with spatial features. Extensive experiments demonstrate that our SAM2Refiner achieves superior segmentation performance on both image and video segmentation tasks, significantly outperforming other state-of-the-art methods.

\clearpage

{\small
\bibliographystyle{ieeenat_fullname}
\bibliography{references}
}

\end{document}